\pdfoutput=1
\documentclass[11pt]{article}

\usepackage[]{EACL2023}

\usepackage{times}
\usepackage{latexsym}

\usepackage[T1]{fontenc}
\usepackage[utf8]{inputenc}

\usepackage{microtype}

\usepackage{inconsolata}

\usepackage{graphicx}
\usepackage{multirow}
\usepackage{multicol}
\usepackage{arydshln}
\usepackage{tikz}
\usepackage{amsmath}
\usepackage{ctable}
\usepackage{adjustbox}
\usepackage{hyperref}
\usepackage{amssymb}
\usepackage{bm}

\usepackage{courier}
\usepackage{pifont}
\usepackage{scrextend}
\usepackage{footmisc}
\usepackage{bbm}

\usepackage{microtype}
\usepackage{multirow}
\usepackage{graphicx}
\usepackage{booktabs}

\usepackage{amssymb}
\usepackage{textcomp}
\usepackage{colortbl}
\usepackage{bibentry}
\usepackage{float}
\usepackage{arydshln}
\usepackage{threeparttable}
\usepackage{mathtools}
\usepackage{bbm}
\usepackage{chngpage}
\usepackage{array}
\usepackage{adjustbox}

\definecolor{mypink1}{rgb}{0.858, 0.188, 0.478}
\definecolor{mypink2}{RGB}{219, 48, 122}
\definecolor{mypink3}{cmyk}{0, 0.7808, 0.4429, 0.1412}
\definecolor{mygray}{gray}{0.6}
\definecolor{mygreen1}{RGB}{46, 139, 87}
\definecolor{mygreen2}{rgb}{0,153,0}
\DeclareMathOperator*{\argmax}{argmax}
\newcommand{\xmark}{\ding{55}}%

\newcommand\sudan[1]{\textcolor{black}{#1}}
\newcommand\mm[1]{\textcolor{black}{#1}}

\title{Context Generation Improves Open Domain Question Answering}

\author{Dan Su$^\ddagger$\thanks{~~This work was done when the first author was an intern at NVIDIA. Corresponding authors: Dan Su, Mostofa Patwary.}, Mostofa Patwary$^\mathsection$, Shrimai Prabhumoye$^\mathsection$, Peng Xu$^\mathsection$, Ryan Prenger$^\mathsection$, \\ \textbf{Mohammad Shoeybi$^\mathsection$, Pascale Fung$^\ddagger$, Anima Anandkumar$^\mathsection$, Bryan Catanzaro$^\mathsection$} \\ 
$^\ddagger$The Hong Kong University of Science and Technology, $^\mathsection$NVIDIA \\
\texttt{dsu@connect.ust.hk}, \texttt{mpatwary@nvidia.com}
}

\begin{document}
\maketitle
\begin{abstract}
% \textit{Closed-book} question answering (QA) requires a model to directly answer an open-domain question without access to any external knowledge. Prior work on \textit{closed-book} QA either directly finetunes or prompts a pretrained language model (LM) to leverage the stored knowledge. However, they do not fully exploit the parameterized knowledge. To address this inefficiency, we propose a two-stage, \textit{closed-book} QA framework which employs a \textit{coarse-to-fine} approach to extract the relevant knowledge and answer a question. We first generate a related context for a given question by prompting a pretrained LM. We then prompt the same LM to generate an answer using the generated context and the question. Additionally, to \mm{reduce} failure caused by context uncertainty, we marginalize over generated contexts. Experimental results on three QA benchmarks show that our method significantly outperforms previous \textit{closed-book} QA methods (e.g. exact matching 68.6\% vs. 55.3\%), and is on par with \textit{open-book} methods that exploit external knowledge sources (e.g. 68.6\% vs. 68.0\%). Our results show that our new methodology is able to better exploit the stored knowledge in pretrained LMs without adding extra learnable parameters or needing finetuning, and paves the way for hybrid models that integrate pretrained LMs with external knowledge.

\textit{Closed-book} question answering (QA) requires a model to directly answer an open-domain question without access to any external knowledge. Prior work on \textit{closed-book} QA either directly finetunes or prompts a pretrained language model (LM) to leverage the stored knowledge. However, they do not fully exploit the parameterized knowledge. To address this inefficiency, we propose a two-stage, \textit{closed-book} QA framework which employs a \textit{coarse-to-fine} approach to extract the relevant knowledge and answer a question. We first generate a related context for a given question by prompting a pretrained LM. We then prompt the same LM to generate an answer using the generated context and the question. Additionally, \mm{we marginalize over the generated contexts to improve the accuracies and reduce context uncertainty}. Experimental results on three QA benchmarks show that our method significantly outperforms previous \textit{closed-book} QA methods. \mm{For example on TriviaQA, our method improves exact match accuracy from 55.3\% to 68.6\%, and is on par with \textit{open-book} QA methods (68.6\% vs. 68.0\%)}. Our results show that our new methodology is able to better exploit the stored knowledge in pretrained LMs without adding extra learnable parameters or needing finetuning, and paves the way for hybrid models that integrate pretrained LMs with external knowledge.
\end{abstract}

\section{Introduction}
\label{sec:introduction}

% Open-domain question answering (ODQA) has been extensively studied in recent years. Significant progress has been made by the \textit{open-book} QA models~\cite{chen2017reading, rag, guu2020realm, izacard2021leveraging, lazaridou2022internet} that explicitly exploit external knowledge corpus via dense retrieval techniques like DPR~\cite{karpukhin2020dense}. However, learning a good retriever requires substantial resources, such as a large number of labeled domain-specific pairs of question and contexts~\cite{karpukhin2020dense}, or intensive computate resources~\cite{lee2019latent}. Also it is more challenging to retrieve relevant knowledge when the size of the external database increases~\cite{reimers-gurevych-2021-curse}. 

\begin{figure}[!t]
 \centering
 \includegraphics[width=0.48\textwidth]{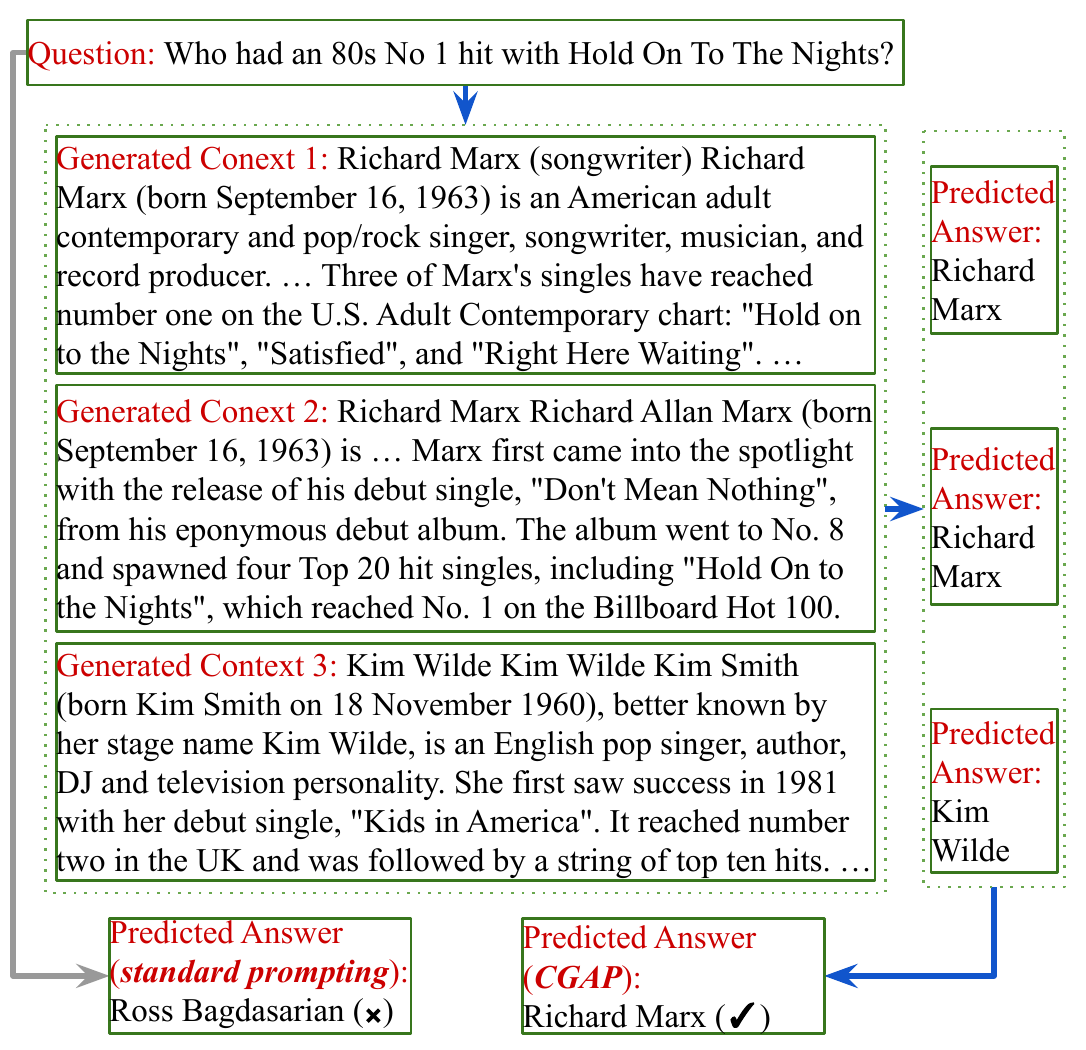}
%  \vspace{-15pt}
  \caption{An example illustrating our two-stage, \textbf{CGAP} framework. CGAP generates more accurate answer (e.g. \textit{Richard Marx}) compared to standard few-shot prompting (e.g. \textit{Ross Bagdasarian}).}
%   \vspace{-15pt}
  \label{fig:example}
\end{figure}

Open-domain question answering (ODQA) \mm{produces an answer to a given question in the form of natural language, and the task} has been extensively studied in recent years. Significant progress on ODQA has been \mm{made by developing } the \textit{open-book} QA methods~\cite{chen2017reading, rag, guu2020realm, izacard2021leveraging, lazaridou2022internet} that explicitly exploit external knowledge corpus via dense retrieval techniques like DPR~\cite{karpukhin2020dense}. However, learning a good retriever requires substantial resources, such as a large number of domain-specific pairs of question and contexts in the knowledge corpus~\cite{karpukhin2020dense}, or intensive compute resources~\cite{lee2019latent}. \mm{In addition, as the size of the knowledge corpus increases, it becomes harder to retrieve accurate contexts due to the high dimensionality of the search space}~\cite{reimers-gurevych-2021-curse}.

% The \textit{open-book} QA methods extract knowledge from an external corpus to answer the question. Significant progress has been made in recent years, with the rapid progress in dense retrieval techniques~\cite{karpukhin2020dense}. 

Another class of models, known as \textit{closed-book} question answering (CBQA), were recently proposed ~\cite{roberts2020much}. CBQA tries to directly answer the open-domain questions without accessing any external knowledge sources, and instead leverages the parametric knowledge stored in the pretrained language models (LMs)~\cite{raffel2020exploring, brown2020language, ye2020studying}. However, even with larger LMs, the \textit{closed-book} methods are not competitive with the \textit{open-book} methods in term of accuracy\mm{~\cite{lewis2021paq}}.

While it has been shown that large pretrained LMs store an abundant amount of knowledge~\cite{petroni2019language,roberts2020much}, we hypothesize the accuracy gaps are largely because the way of exploiting the parameterized knowledge are not sophisticated enough. Prior works on CBQA either finetune pretrained LM models on the entire QA datasets~\cite{roberts2020much, ye2020studying}, or they directly prompt those models using several few-shot QA pairs~\cite{brown2020language, radford2019language}. On the contrary, \textit{open-book} models use a two-stage pipeline. They first retrieve relevant contexts from external corpus, then they extract the answer based on the retrieved contexts.

%\mm{Prior works on CBQA finetune pretrained LMs on the entire QA datasets~\cite{roberts2020much, ye2020studying}. However, the fine-tuning based method has limitations to generalize to new domains~\cite{talmor2019multiqa, su-etal-2019-generalizing,liu2021challenges}, also it is hard to fine-tune large pretrained LMs such as GPT-3~\cite{brown2020language}, Megatron-Turing-530B~\cite{smith2022using}, and PALM~\cite{chowdhery2022palm} due to smaller finetuning datasets. Other work directly prompt pretrained LMs using several few-shot QA pairs~\cite{brown2020language, radford2019language} in a single stage. While \textit{open-book} models use a two-stage pipeline, we hypothesis that a two-stage prompting will exploit the knowledge in pretrained LMs better.} 

% or they  On the contrary, \textit{open-book} models use a two-stage pipeline. They first retrieve relevant contexts from external corpus, then they extract the answer based on the retrieved contexts.

% While it has been shown that large pretrained LMs store an abundant amount of knowledge~\cite{petroni2019language,roberts2020much}, 

Therefore, to better exploit the parameterized knowledge in pretrained LMs and bridge the large accuracy gaps between the \textit{closed-book} and \textit{open-book} methods, we propose a \textit{coarse-to-fine}, two-stage method for CBQA task. The main idea is to leverage generated contexts as an intermediate bridge between the huge amount of parameterized knowledge stored in the LM and the answer that lies within this knowledge. To the best of our knowledge, no previous work has been conducted on generating context from large pretrained LMs for CBQA and leveraging them to predict answer.

Our proposed framework \textbf{CGAP} consists of two stages. It first performs \textbf{C}ontext \textbf{G}eneration relevant to a given question by prompting a pretrained LM. It then prompts the same LM for \textbf{A}nswer \textbf{P}rediction using the generated context and the question.  
In order to improve the accuracies and to reduce context uncertainties, we generate multiple contexts for each question and predict the final answer by majority voting. This step does not increase the inference cost as we generate the contexts in parallel by batching in a single inference call.
%In order to mitigate failures caused by variability in the generated context, we generate multiple contexts, predicting the final answer by majority voting. 
Figure~\ref{fig:example} illustrates how our two stage prompting and majority voting works. For the input question, CGAP generates 3 contexts and 3 predicted answers at the two stages respectively, and choose the most voted answer as the final answer. 
Note that we do not finetune the large pretrained LMs for context generation or answer prediction. This facilitates our approach to take advantage of large LMs such as GPT-3~\cite{brown2020language}, PALM~\cite{chowdhery2022palm}  or Megatron-Turing NLG 530B~\cite{smith2022using}, which are only available through APIs.

%Our proposed framework \textbf{CGAP} consists of two stages. It first performs \textbf{C}ontext \textbf{G}eneration relevant to a given question by prompting a pretrained LM. It then prompts the same LM for \textbf{A}nswer \textbf{P}rediction using the generated context and the question.  In order to mitigate failures caused by variability in the generated context, we generate multiple contexts, predicting the final answer by majority voting. \mm{The multiple contexts can be generated in parallel by batching and do not impose extra latency}. Figure~\ref{fig:example} \mm{illustrates how our two-stage prompting and majority voting works. For the input question, CGAP generates 3 contexts and 3 predicted answers at the two stages respectively, and choose the most voted answer as the final answer}. Note that we do not finetune the large pretrained LMs for context generation or answer prediction. This facilitates our approach to take advantage of large LMs such as GPT-3~\cite{brown2020language}, PALM~\cite{chowdhery2022palm}  or Megatron-Turing NLG 530B~\cite{smith2022using}, which are only available through APIs.

We conduct in-depth experimental studies on three open-domain QA benchmarks, Natural Questions~\cite{kwiatkowski2019natural},
WebQuestions~\cite{berant2013semantic}, and TriviaQA~\cite{joshi2017triviaqa}, and demonstrate significant improvements by our two stage prompting method. Our contributions are summarized as follows:

\begin{itemize}
\setlength{\itemsep}{3pt}
\setlength{\parsep}{0pt}
\setlength{\parskip}{0pt}
    \item We propose a simple yet effective few-shot prompting approach for ODQA that does not rely on any external knowledge sources or fine-tuning, but performs significantly better than existing \textit{closed-book} approaches (e.g. exact matching 68.6\% vs. 55.3\%), and is on par with \textit{open-book} methods (e.g. 68.6\% vs. 68.0\%). 
    % \item To the best of our knowledge, we are the first to leverage generated context from large pretrained LMs for open-domain question answering.
    % \item We show that the generated context can improve standard few-shot prompting based \textit{closed-book} QA accuracy at various model scales (e.g. from 11.7\% to 28.5\%), and scaling up the context generation model further enlarges their accuracy gaps (e.g. from 28.5\% to 36.3\%).
    \item We show that the generated context can improve standard few-shot prompting based
    \textit{closed-book} QA accuracy at various model scales (e.g. from 11.7\% to 28.5\%), and demonstrate that scaling up the context generation model further enlarges their accuracy gaps \mm{(e.g. 357M 28.5\% vs. 530B 68.6\%)}. \mm{To the best of our knowledge, we are the first to leverage generated context from large pretrained LMs for open-domain question answering.}
    
    \item We show that generating multiple contexts without increasing the inference cost by batching can mitigate errors in answer prediction caused by variability in the unknown context (e.g. from 36.3\% to 45.7\%). 
    %\item \mm{We propose multiple contexts generation that can be batched with minimal cost addition and show} that generating multiple contexts can mitigate errors in answer prediction caused by variability in the unknown context (e.g. from 36.3\% to 45.7\%).  
    % \item We scale our method up to 530B parameter models and showcase that larger models boost the accuracy by huge margins (e.g. 357M 28.5\% vs. 530B 68.6\%).
\end{itemize}

% ADD a paragraph like outlining the structures.

\section{Methodology}
\label{sec:methodology}

\begin{figure*}[!ht]
 \centering
 \includegraphics[width=0.94\linewidth]{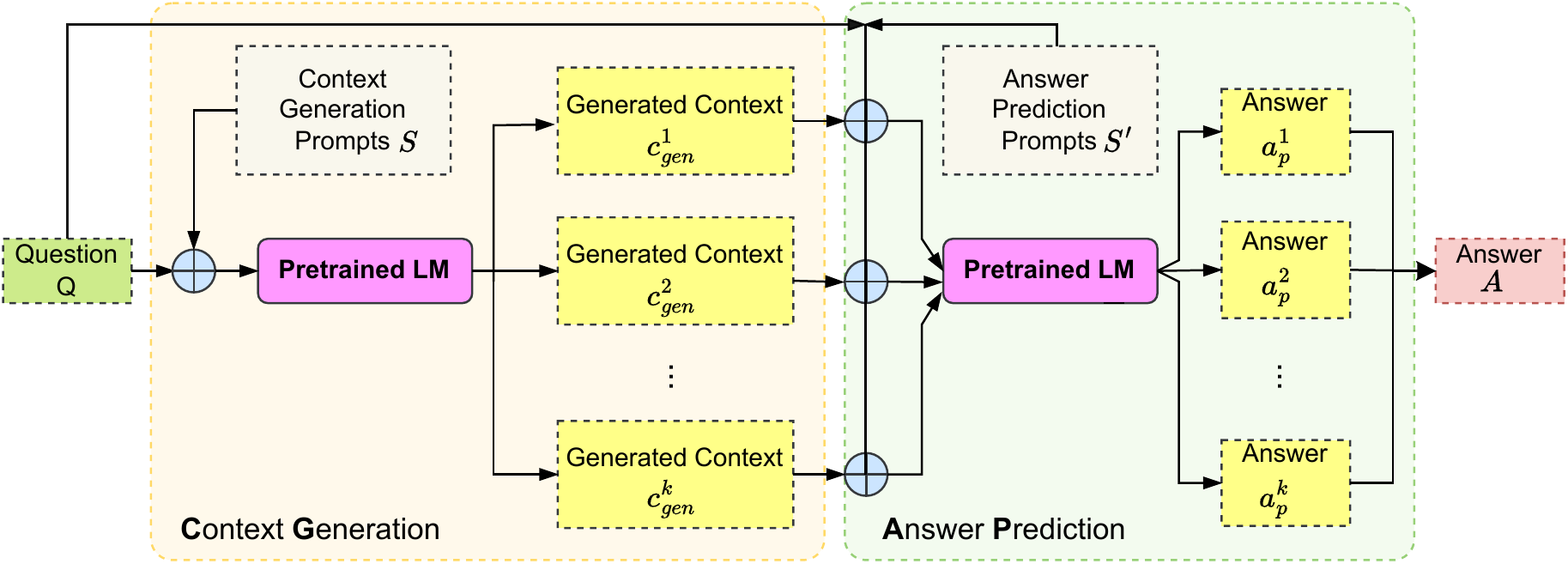}
  \caption{Overview architecture of our \textbf{CGAP} framework. It first does \textbf{C}ontext \textbf{G}eneration by prompting large pretrained LMs, then it further prompts the LMs for \textbf{A}nswer \textbf{P}rediction by feeding the generated context to the LM models alongside the question. $k$ contexts are generated and the final answer $A$ is chosen by majority voting. (\textit{If computation capability allows, it could prompt multiple ($k$) LMs in parallel at both two stages to speed up.}) }
  \label{fig:framework}
\end{figure*}

Our proposed \textbf{C}ontext \textbf{G}eration and \textbf{A}nswer \textbf{P}rediction (\textbf{CGAP}) framework is illustrated in Figure~\ref{fig:framework}. CGAP consists of two stages. First, it generates relevant context to a given question by prompting a large pretrained LM. 
In the second stage, it predicts an answer using the generated context and the question by prompting the same LM. 
To accurately predict the answer, we generate multiple contexts. We run each of the two stages multiple times in parallel in batch for the same question, generating different contexts for each, and use majority voting to select the final answer.

% Large LMs such as GPT-3~\cite{brown2020language}, Megatron-Turing-530B~\cite{smith2022using}, and PALM~\cite{chowdhery2022palm} have shown promising results in few-shot learning. They have also been shown as a good knowledge source~\cite{roberts2020much, liu2022multi}. However, finetuning them on downstream applications is challenging due to smaller datasets. Moreover, they are only available through API. In our \textbf{CGAP} framework, we therefore propose a modified few-shot prompting and demonstrate significant improvement over exiting finetuing or few-shot accuracies. 

% We denote the input question as $Q$, and the corresponding generated context as $C_{gen}$. We use  $C_{gen}^i$ to represent the $i$-th generated context samples that we marginalize on $(i=1.,,,k)$. The database of samples is denoted as $D$, and each data sample in $D$ is denoted by $d_i$ where $i=1,...N$, which consists N triples $<Q_i, C_i, A_i>$ , a question $Q_i$, a golden context passage $C_i$, and corresponding answer $A_i$.

Formally, for our task we have a question $Q$ to be answered, and a support repository $\mathcal{D} = \{(c_1, q_1, a_1), \ldots, (c_n, q_n, a_n)\}$ that consists of tuples of question $q_i$ and answer $a_i$ pairs with mapping to the context $c_i$.
\sudan{In our experiments, we use the training sets of the corresponding datasets as $\mathcal{D}$.}

\subsection{Context Generation}
\label{sec:context_generation}

% We would like to leverage the parameterized knowledge stored in large pretrained LMs to facilitate the answer generation process.
% We first prompt the LM with few-shot examples to generate contexts that are relevant to the given question.

\mm{As shown in Figure~\ref{fig:framework}, in the first stage, given question $Q$, we select the $m$ context generation prompts $S = \{(q_1, c_1), \ldots, (q_m, c_m)\}$ from the support repository $\mathcal{D}$.
We then use $S$ with $Q$ to prompt pretrained LM to generate $k$ contexts, which are denoted by $ C_{gen} = \{c^1_{gen}, c^2_{gen}, \ldots, c^k_{gen}\}$.}

\paragraph{Sample Selection} Selecting appropriate samples \mm{for} the prompts is the key to generate high-quality context relevant to \mm{a given question}. 
Previous work has shown that leveraging relevant samples helps the LM to generate contextually relevant and factually correct context~\cite{liu2021pre,liu2022multi}. We therefore use a similarity-based retriever to search relevant samples \mm{$S$} from the corresponding \sudan{supporting repository, $\mathcal{D}$}. We use DPR~\cite{karpukhin2020dense} in our framework. In our DPR setup, we represent the question and the samples in \sudan{$\mathcal{D}$} as $768$-dimensional dense vector representations, computed via the BERT-based bi-encoder networks. We rank the documents according to their similarity score, calculated as:
\begin{equation}
        Score(Q, (q_j, c_j)) = \text{BERT}(Q)^T \cdot \text{BERT}(q_j; c_j)
\label{eqa:retriever_score}
\end{equation}

where $;$ denotes concatenation of the tokens of the question $q_j$ and the context $c_j$. 
Finally, we get $S=\{(q_1, c_1), \ldots, (q_m, c_m)\}$ which are the top-m retrieved samples for question $Q$.

\sudan{We would like to emphasize that the selected samples from $\mathcal{D}$ are used as examples in the few-shot prompting to the pretrained LM to generate context, not as the source of external knowledge containing the answer. } 
% We do not use the training dataset or those selected samples for finetuning or additional training to answer the question, rather we rely on the knowledge stored in the pretrained parameters in LM.
% }

%\mm{Also, they do not include the knowledge required to predict the answers.}

%our method does not use the supporting repository as an external knowledge base, rather we use few of them as examples to the pretrained LM to generate context.}
% feeding the pretrained LM with suitable and intuitive prompts can trigger it to generate relevant content
\paragraph{Prompts Construction} Given the question $Q$ and the set of question-context pair samples $S$ selected, we use few-shot prompting to condition pretrained LMs on the samples. 
We use similar few-shot prompting technique for \textit{closed-book} QA as in ~\cite{brown2020language}, that considers multiple <question, answer> pairs. The template we used to construct prompts is: \texttt{Q:} ... \texttt{A:} .... 
Thus the constructed prompt $Prompt(Q)$ for a given question $Q$ becomes:
\begin{align*}
    Prompt(Q) =& \texttt{Q:} q_m \backslash n \texttt{A:} c_m \backslash n \ldots \\ &\texttt{Q:} q_1 \backslash n \texttt{A:} c_1 \backslash n \texttt{Q:} Q \backslash n
\end{align*}

We use '$\backslash n$' to separate the question, context and the samples. 
We investigated the order of samples to optimize the prompt and find that using the retrieved samples in reversed order of similarity yields better accuracies across all datasets\footnote{\sudan{We show an concrete example of $Prompt(Q)$ in Appendix Table \ref{tab:appendix-promptq}}}. 
We now pass $Prompt(Q)$ through a pretrained LM to generate the context as follows: 

\begin{equation*}
    c_{gen} = \mathcal{LM}(Prompt(Q))
\end{equation*}

% To generate a set of $k$ contexts, $\{c^1_{gen}, ..., c^k_{gen}\}$, we run this step $k$ times.
\mm{To generate a set of $k$ contexts, $\{c^1_{gen}, ..., c^k_{gen}\}$, we increase the inference batch size to $k$ and generate all the $k$ contexts in parallel in one inference call to the LM. Thus, the overall latency remains the same as using a single context.}

\subsection{Answer Prediction}
\label{sec:ans_pred}
\mm{In the second stage, we select $m$ answer prediction prompts $S'=\{(q_1, a_1, c_1), \ldots, (q_m, a_m, c_m)\}$ from $\mathcal{D}$ and then we prompt the same LM using the generated context $C_{gen}$ from the first stage, along with the question $Q$ and $S'$. The LM predicts a set of $k$ answers $A_p = \{a^1_p, a^2_p,..., a^k_p\}$ each corresponding to the $k$ contexts in $C_{gen}$. The final answer $A$ is selected by majority voting on $A_p$.}

% In the answer prediction stage, we use the generated context $c_{gen}$ from the first stage along with the question $Q$. Specifically, we prompt the same LM by few-shot examples selected from the \sudan{supporting repository}, $\mathcal{D}$, together with $c_{gen}$ and $Q$. 

\paragraph{Sample Selection} 
Constrained by the maximum sequence length of the LM, we can feed the LM only a few $(c, q, a)$ samples. 
Thus, it could be difficult for the LM to learn how to predict the answer for the given question conditioned on the context, unless similar examples have been provided. 
For example, if we were asking the question \textit{'who is the current director of the us mint?}', the example that answering the question \textit{'who is the fbi director of the united states?'} from the provided context will be more helpful, than the example that is answering \textit{'how many episodes are there in `Dragon Ball Z'?'} from the given context. We therefore use the same criteria for answer prediction as has been used for context generation. 
We use the same set of samples as selected in the first stage as described in Equation~\ref{eqa:retriever_score} and denote as $S'=\{(q_1, c_1, a_1), \ldots, (q_m, c_m, a_m)\}$. %However, each $S'_i$ is composed of tuples $<Q_i, C_i, A_i>$, where $A_i$ is the corresponding answer for question $Q_i$ and given context $C_i$. 
%Thus, we use the same similarity based criteria to select the answer prediction prompt samples, i.e., we used the same set of samples selected from the first stage as described in Equation~\ref{eqa:retriever_score}, we denote as $S'=\{S'_1, S'_2,..., S'_m\}$, except that now $S'_m$ is composed of triples of $<Q_m, C_m, A_m>$ where $A_m$ is the corresponding answer for question $Q_m$ given context $C_m$. 

\paragraph{Prompt Construction} We are prompting LMs with few-shot examples to predict answer for the question conditioned on the generated context. 
To equip the LM with this capability, we constructed intuitive prompts for the selected examples and feed them into the LM. 
Specifically, the template we used to construct answer prediction prompts is: \texttt{C:} ... \texttt{Q:} ... \texttt{A:} ... . 
Thus, the constructed prompt for a given question Q and the $i$-th generated context $c^i_{gen}$ is:
\vspace{-5pt}

\begin{align}
\begin{split}
    Prompt(c^i_{gen}, Q) =&
    \texttt{C:} c_m \backslash n 
    \texttt{Q:} q_m \backslash n
    \texttt{A:} a_m \backslash n\\
    & \ldots \\
    &\texttt{C:} c_1 \backslash n 
    \texttt{Q:} q_1 \backslash n 
     \texttt{A:} a_1 \backslash n\\
    &\texttt{C:} c^i_{gen} \backslash n
    \texttt{Q:} Q \backslash n
    \label{equ:cgen_q}
\end{split}
\end{align}

We then feed $Prompt(c^i_{gen}, Q)$ into the pretrained LM to predict the answer:
\begin{equation}
    a_p^i = \mathcal{LM}(Prompt(c^i_{gen}, Q)))
    \label{equ:answer_prediciton}
\end{equation}
where we use $a^i_p$ to denote the $i$-th answer predicted by the LM. The $k$ generated contexts in $c_{gen}$ will yield a set of answers $A_p=\{a^1_p, ...,a^k_p\}$.

\subsection{Context Marginalization}
\label{sec:methods_margin}
The large pretrained LM can generate impressively fluent and relevant context given input, it also has a tendency to generate factually incorrect statements, ranging from subtle inaccuracies to wild hallucinations~\cite{shuster2021retrieval, krishna2021hurdles, su2022read}. 
Answers conditioned solely on hallucinated or erroneous statements are likely to be incorrect (Equation~\ref{equ:answer_prediciton}). Thus, we would like to remove the variability in the answer due to any particular generated context.

Ideally, we could marginalize over this unknown context by producing an answer for every possible context, weighting each answer by the probability of the context. Here we approximate this by generating a set of contexts, and selecting the final answer based on majority voting. Suppose there are $T$ unique answers $\{A^1_p,...,A^T_p\}$ from the $k$ predicted answer from Equation~
\ref{equ:answer_prediciton} where $T<=k$, then we select the $J$-th answer that receives the highest number of votes from the $T$ different answers via:
\begin{equation}
    J = \argmax_{j\in \{1,2,...,T\}} \sum^k_{i=1}(\mathbbm{1}(a^i_p = A^j_p))
    \label{equ:marginalization}
\end{equation}

as the final answer $A$.  As $k$ gets larger, the final answer $A$ will converge to the answer that would be produced marginalizing over all possible contexts.  We refer to this majority vote over multiple generated contexts as context marginalization.

% \sudan{Notably, even though we generate $k$ contexts to answer a question, increasing the batch size to $k$ can generate all the $k$ contexts in parallel in one inference call to the LM. Thus, the overall latency can remain the same as using a single context.}

\begin{table*}[!th]
\centering
\begin{adjustbox}{width=0.95\textwidth}
{
\begin{tabular}{lllccc}
\specialrule{.08em}{.1em}{.1em} 
\textbf{Model Type}  & \multicolumn{1}{c}{\textbf{Model}} & \textbf{Method} &\textbf{NQ} & \textbf{TQA} & \textbf{WQ}\\ 
\hline 
\multirow{3}{*}{Open-book} & RAG~\cite{lewis2020retrieval} & \textit{Finetuned} & 44.5  & 68.0    & \textbf{45.5}  \\
                            & Fusion-in-Decoder (large) ~\cite{izacard2021leveraging} & \textit{Finetuned}  & \textbf{51.4}  & 67.6  & -     \\
                              & $OB_{Google}^{PoE}$~\cite{lazaridou2022internet} & \textit{Few-shot} & 38.4  & - & - \\ 
                              \hline
\multirow{5}{*}{Closed-book}  & T5-11B ~\cite{roberts2020much} & \textit{Finetuned} & 32.6             & 42.3     & 37.2         \\
                              & T5-11B+SSM ~\cite{roberts2020much} & \textit{Finetuned}   & 34.8  & 51.0    & 40.8  \\
                            & BART-large, pre-finetuned on PAQ ~\cite{lewis2021paq} & \textit{Finetuned} & 32.7  & 33.2  & -  \\
                            %   & GPT-3-175B (Few-shot, paper) *~\cite{brown2020language}              & 29.9  & 71.2  & 41.5  \\
                              & LM-530B (API) & \textit{Few-shot}           & 23.0 & 55.3 & 23.6 \\ 
                              \cdashline{2-6}
                            % \cline[dashed]{2-6}
                              & \textbf{CGAP} (ours, 530B)   & \textit{Few-shot} & \underline{42.0} & \textbf{\underline{68.6}}  & \underline{41.8} \\ 
\specialrule{.08em}{.1em}{.1em} \\
\end{tabular}
}
\end{adjustbox}
% \vspace{-10pt}
\caption{Exact Match score for \textbf{CGAP} (highest accuracy configurations) in comparison to recent state-of-the-art \textit{open-book} and \textit{closed-book} based systems. Highest score indicated in \textbf{bold}, highest \textit{closed-book} model
\underline{underlined}.}
\vspace{-5pt}
\label{tab:main_results}
\end{table*}

\section{Experimental Setup}

\subsection{Datasets}
\label{sec:datasets}
We evaluated our experiments on three open-domain QA benchmark datasets: Natural Questions (NQ)~\cite{kwiatkowski2019natural}, TriviaQA (TQA)~\cite{joshi2017triviaqa}, and WebQuestions 
(WQ)~\cite{berant2013semantic}, using the same data splits for train, validation and test as in~\citet{lee2019latent, izacard2021leveraging}.

NQ contains questions from Google search queries; TQA contains a collection of questions from trivia and quiz-league websites, and we use their unfiltered set; while questions of WQ were from Google Suggest API. 
For NQ and TQA, we use the processed data provided by~\citet{izacard2021leveraging}, in which each question-answer pair is accompanied by a 100-words Wikipedia passage containing the answer. 
For WQ, we retrieved the corresponding context passage for each question from 2019/08/01 Wikipedia dump, using the DPR-based retriever that is trained jointly on the union of knowledge-intensive training data in KILT benchmark~\cite{petroni2021kilt}.

%WE NEED TO MENTION ABOUT THE PRETRAINED MODELS as well.

\subsection{Baselines}
\label{sec:baselines}
We compare our \textbf{CGAP} framework with the following baseline methods for \textit{closed-book} QA.
\vspace{-5pt}
\paragraph{Standard Few-shot Prompting} 
We use the standard few-shot prompting technique
similar to GPT-3~\cite{brown2020language} in our evaluation on the \textit{closed-book} QA datasets as described in Section ~\ref{sec:datasets}. We consider this technique as the few-shot baseline in all our experiments. The baseline that is experimented using 530 billion (530B) parameterized LM is refferred as \textbf{LM-530B}.
%For fair comparison of this technqiue with our method, we experiment the same 530 billion parameterized LM (referred as \textbf{LM-530B}).
%adopted the standard few-shot prompting on GPT-3 for \textit{closed-book} QA, and evaluated on the same datasets as described in Section ~\ref{sec:datasets}.
%However, we were unable to reproduce the results reported in their paper when we query GPT-3 via API~\footnote{Results of querying OpenAI GPT-3 API using standard prompting are shown in Appendix~\ref{sec:appendix-gpt3}}. 
%Thus, we use their standard prompting format and experimented on a 530 billion parameterized LM as our baseline (referred as \textbf{LM-530B}), for fair comparison. 
%Greedy

% They use beam search with a beam width of 4 and a length penalty of 0.6 for answer prediction. (reffered as \textbf{GPT-3 (Few-shot, paper)})
% They first randomly draw $K$ (=64) question-answer pairs from the corresponding training set, and use 'Q: ' and 'A: ' respectively as prefix before each question and answer, to form the conditioning prompts.

\paragraph{LM Fune-tuning}~\citet{roberts2020much} first proposed the \textit{closed-book} QA task for open domain QA, and they directly fine-tuned T5~\cite{raffel2019exploring} using the entire QA pairs in the training data, without access to any external knowledge corpus (referred as \textbf{T5-11B}). They also experimented with using ’Salient Span-Masking’ (SSM) to continue pretraining the T5 checkpoints before fine-tuning for QA (referred as \textbf{T5-11B+SSM}). ~\citet{lewis2021paq} pre-finetuned BART-large~\cite{lewis2020bart} on \textit{Probably Asked Questions} (PAQ), a very large resource of 65M automatically generated QA-pairs, then further finetuned the model on corresponding training data (referred as \textbf{BART-large, pre-finetuned on PAQ}).

\boldmath
\paragraph{Open-book Few-shot Prompting} ~\citet{lazaridou2022internet} used few-shot prompting for open domain QA task, but they generate the answer via conditioning on retrieved documents from Google Search API. (referred as \textbf{$OB^{PoE}
_{Google}$})
% They proposed to do answer ranking via Product-of-Experts(PoE) to combine together all probabilities for the final answer probability. 
\unboldmath

\subsection{State-of-the-art Open-book QA Models}
We compare the state-of-the-art \textit{open-book} QA models with \textbf{CGAP}. \textbf{Fusion-in-Decoder} (FiD)~\cite{izacard2021leveraging} uses DPR~\cite{karpukhin2020dense} to retrieve 100 passages from Wikipedia. Then they encode each passage independently and combine all outputs from the T5 encoder before passing them to the T5 decoder to generate a final answer. \textbf{RAG}~\cite{rag} is an end-to-end retrieval-augmented generation model. 
% which back-propagates to the retriever’s input encoder, learning to adapt the input embedding to retrieve more relevant results.

% \begin{figure*}[!t]
% 	\begin{center}
% 		\includegraphics[width=0.95\textwidth]{figs/ablation-1-new.pdf}
% 		\caption{Ablation on context generation LM Size. Answer prediction LM size: 357M (\textit{left}), 1.3B(\textit{right}). The colored dash lines represent the standard prompting few-shot baselines.}
% 		\vspace{-5pt}
% 		\label{Fig:ablation-CG-357m-1.3b-ans}
% 	\end{center}
% \end{figure*}

\begin{figure*}[!t]
	\begin{center}
		\includegraphics[width=0.95\textwidth]{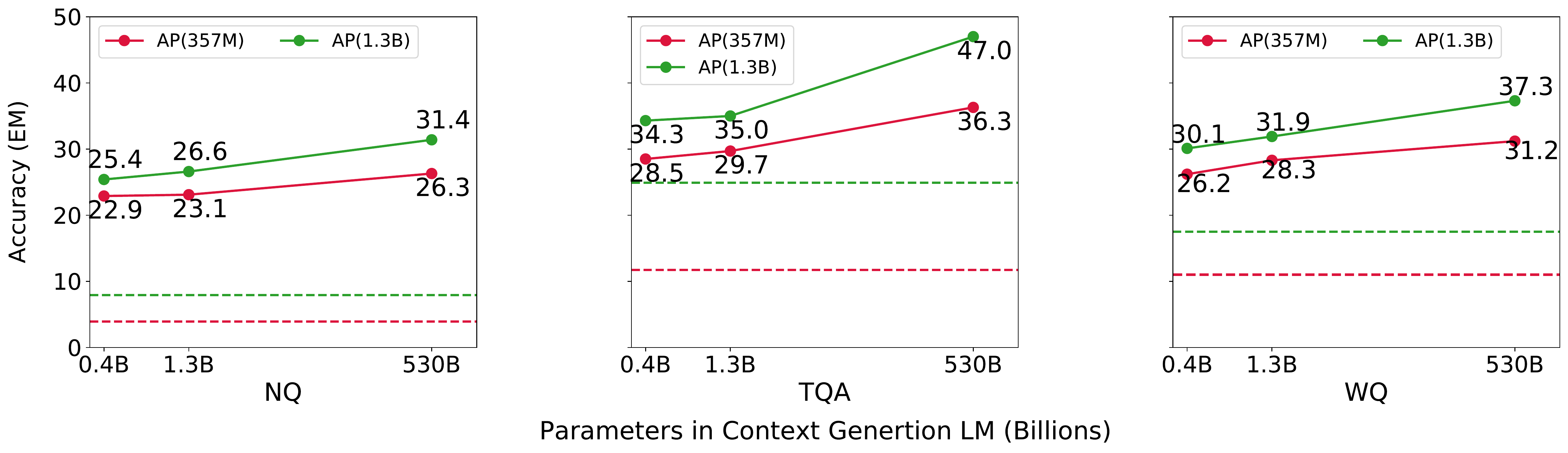}
		\caption{Ablation on context generation LM size. The dash lines represent standard few-shot prompting baselines.}

		\label{Fig:ablation-CG-357m-1.3b-ans}
	\end{center}
\end{figure*}

\subsection{Implementation Details}
To test how different model scales affect the performance of our approach, \sudan{we train and experiment on a collection of dcoder-only LMs using the Megatron-LM framework~\cite{shoeybi2019megatron}}, with 357 million (357m),
1.3 billion (1.3b), and 530 billion (530b)~\cite{smith2022using} parameters, at both context generator and answer prediction stage. 
We use top-$p$ sampling with a value of 0.9 to generate diversified contexts. However, to handle the deterministic generation (e.g. short answer),
we use greedy decoding at the answer prediction stage, similar to ~\cite{chowdhery2022palm, wang2022self}.

For the prompt configuration at both stages, we choose 10 samples, constrained by the maximum sequence length of the LMs. We use DPR checkpoint from Huggingface\footnote{\url{https://huggingface.co/facebook/dpr-ctx_encoder-multiset-base}} to select samples from the \sudan{supporting repository}.

\subsection{Evaluation}
% Following the standard evaluation procedures in previous work~\cite{rajpurkar2016squad,lee2019latent, izacard2021leveraging}, we use Exact Match (EM) as our answer accuracy evaluation metric, where each predicted answer is compared to the ground-truth after both are lowercased and stripped of articles, punctuation, and duplicate whitespace.

\sudan{For evaluating the open-domain QA task, we followed the recent works~\cite{rajpurkar2016squad,lee2019latent, izacard2021leveraging} that use Exact Match (EM) as the evaluation metric}. Each predicted answer is compared to the ground-truth after both are lowercased and stripped of articles, punctuation, and duplicate whitespace.

\section{Results and Ablation Studies}
We now show our main results as well as ablations to further analyze the effectiveness of our approach.
% ompare our CCAP approach with baselines, and also state-of-the-art \textit{open-book} models for open-domain QA. 

\begin{table*}[t]
\centering
% \resizebox{.5\textwidth}{!}{
\begin{adjustbox}{totalheight=0.21\textheight-2\baselineskip,}
{
\begin{tabular}{ccclll}
\specialrule{.08em}{.1em}{.1em} 
\begin{tabular}[c]{@{}c@{}}\textbf{AP} \\ \textbf{LM Size}\end{tabular} &
  \begin{tabular}[c]{@{}c@{}}\textbf{CG} \\ \textbf{LM Size}\end{tabular} &
  \begin{tabular}[c]{@{}c@{}}\textbf{Margin-} \\\textbf{alization} \end{tabular} &
  \textbf{NQ} &
  \textbf{TQA} &
  \textbf{WQ} \\ \hline
\multirow{6}{*}{357M} & \multirow{2}{*}{357M} & \xmark & 22.9 & 28.5 & 26.2 \\
                      &  &  \checkmark   & 25.7 \textcolor{mypink2}{\small{(+2.8)}} & 33.4 \textcolor{mypink2} {\small{(+4.9)}} & 29.6 \textcolor{mypink2}{\small{(+3.4)}} \\ \cline{2-6} 
                      & \multirow{2}{*}{1.3B} & \xmark & 23.1 & 29.7 & 28.3  \\
                      &  &  \checkmark  & 26.1 \textcolor{mypink2}{\small{(+3.0)}} & 34.8 \textcolor{mypink2}{\small{(+5.1)}} & 31.3 \textcolor{mypink2}{\small{(+3.0)} } \\ \cline{2-6} 
                      & \multirow{2}{*}{530B} & \xmark & 26.3 & 36.3 & 31.2  \\
                      &  &  \checkmark   & 28.9 \textcolor{mypink2}{\small{(+2.7)}} & 45.7 \textcolor{mypink2}{\small{(+9.4)}} & 34.0 \textcolor{mypink2}{\small{(+2.8)}} \\ \hline
\multirow{2}{*}{530B} & \multirow{2}{*}{530B}  & \xmark & 29.5 & 56.3 & 28.3  \\
                      &  & \checkmark    & \textbf{42.0} \textcolor{mypink2}{\small{(+12.5)}} & \textbf{68.6} \textcolor{mypink2}{\small{(+12.4)}} & \textbf{41.8} \textcolor{mypink2}{\small{(+13.5)}} \\ 
\specialrule{.08em}{.1em}{.1em} 
\end{tabular}
}
\end{adjustbox}{}
\caption{Ablation on context marginalization. (AP and GP represent Answer Prediction and Context Generation, respectively.)}

\label{tab:ablation-2-margin}
\end{table*}

% \subsection{Comparison to Prior Work}
\subsection{Main Results}
\label{sec:main_results}
Table~\ref{tab:main_results} shows the EM score comparison between our CGAP-based method with existing \textit{closed-book} baseline approaches~\footnote{GPT-3 API shows different results than reported in the paper~\cite{brown2020language}. We therefore did not compare to it. 
%We were unable to reproduce the results reported in paper~\cite{brown2020language} when we query GPT-3 via API, so we didn't compare to it. 
Details are shown in Appendix~\ref{sec:appendix-gpt3}}. 
We also compare with state-of-the-art \textit{open-book} models at the upper section of the table. 

% While GPT-3 (175B) few-shot method reported the highest answer accuracy on TQA~\cite{brown2020language}, we can not reproduce that score when we call the GPT-3 API by ourselves~\footnote{Results of querying openAI GPT-3 API using standard prompting are shown in Appendix}. 
As we can see, our CGAP based method outperforms other existing \textit{closed-book} methods by large margin, especially on NQ and TQA datasets. 
The CGAP also outperforms the standard few-shot prompting baseline LM-530B on all three datasets (at least by $13.3$ EM point). 

Furthermore, CGAP obtains highest score on TriviaQA. 
The scores are also very close to the state-of-the-art \textit{open-book} method RAG on NQ and WebQuestions, but only lose few points on NQ to FiD. While FiD uses 100 retrieved passages for answer prediction, CGAP only uses $8$ generated contexts for approximate context marginalization.

\subsection{Ablation Studies}
We conducted a systematic ablation study to further investigate the contribution of the context generation model and the effect of context marginalization.

\subsubsection{Context Generation}
\label{sec: ablation_cg}
While previous work~\cite{roberts2020much, brown2020language} demonstrated that the scale of the model sizes improves the answer accuracy of \textit{closed-book} QA, there are also other findings showing that simply increasing the model size does not lead to substantive accuracy gains~\cite{RaeGopher2022}. Thus, we intend to investigate \textbf{how will the context generation LM affect the answer accuracy}. 

We experimented by varying the LM sizes for context generation, and fix the answer generation LM. We used context generation LM sizes of 357m, 1.3B and 530B, and answer generation LM with 357m and 1.3B parameters. 
We also compare with standard few-shot prompting which has no context generation.
 
We plot the results in Figure~\ref{Fig:ablation-CG-357m-1.3b-ans}. As we can see, there are huge accuracy gains from standard prompting, to CGAP method that has context generation. The accuracy increases by absolute $19.00$\% for NQ, $16.87$\% for TQA and $15.26$\% for WQ, when using 357M model for both standard prompting and CGAP approach.
%For example, when we use the same answer prediction LM (357m), the answer accuracy increases from 3.85\% to 22.85\% for NQ, 10.97\% to 26.23\% for WebQuesions, and 11.66\% to 28.53\% for TriviaQA respectively, when we use only a 357M LM for context generation in CGAP. 
The answer accuracy continues to increase when we increase the LM size for context generation. 
Furthermore, we notice that the slopes of the accuracy gain curve using larger answer prediction model is steeper than using smaller one on all three datasets. This suggests the use of larger answer prediction LM to fully exploit the knowledge in generated context.   

\begin{figure}[!th]
	\begin{center}	\includegraphics[width=0.34\textwidth]{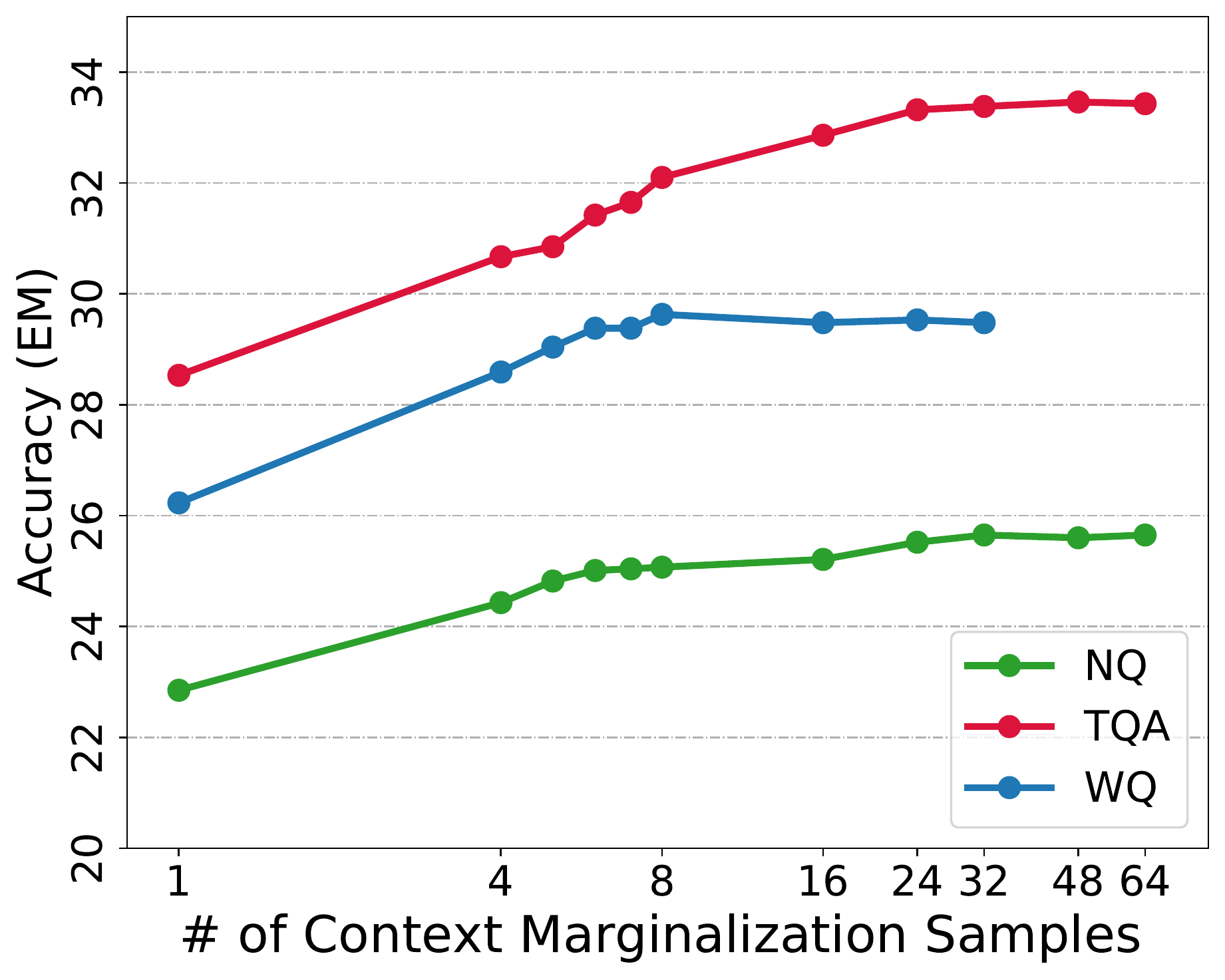}
		\caption{Ablation on $k$, the number of contexts for marginalization.}
		
		\label{Fig:ablation-margin-k}
	\end{center}
\end{figure}

\subsubsection{Context Marginalization}
\label{sec:margin}
Since there will be some hallucinated content or erroneous statements in the generated context, we approximate context marginalization by sampling multiple contexts and selecting the final answer based on majority voting, as introduces in Section~\ref{sec:methods_margin}. Here, we investigate the performance gains brought in by context marginalization, and also the accuracy curves with varied number of sampled contexts used in the approximate marginalization $k$.

In Table~\ref{tab:ablation-2-margin}, we show the accuracy comparisons w/ and w/o using marginalization (k=8), with different LM sizes. 
As we can see, \textbf{context marginalization improves the answer accuracy consistently on the three datasets}\footnote{We show a concrete example in Appendix~\ref{sec:appendix_example} Table~\ref{tab:appendix-gcap}}, under all settings. 
Notably, there is much larger performance gains using marginalization when we scale up the model sizes to 530 billion parameters (i.e. increase EM score by 12.8\% averaged on three datasets). 

The larger the number of context samples $k$, the more accurately the majority vote reflects the true marginalization over all possible contexts.  Therefore, we perform further ablation by changing the value $k$ for 357M LM for both context generation and answer prediction. 
%We use 357M LM , but vary the number of context samples $k$ used for marginalization. 
We plot the accuracy curves in Figure~\ref{Fig:ablation-margin-k}. We see that there are accuracy improvements when we use more context samples.  As expected and curves plateau for larger values of $k$ as the approximation approaches the true marginalization over all possible contexts.

\section{Analysis}
Considering that it is the first time leveraging context generated by large pretrained LMs for ODQA, we also conducted further analysis. 

We compare generated context with retrieved context in the two-stage, few-shot prompting based CBQA framework. It is a dominant paradigm to use retrieved context from external corpus together with the question for answer prediction for \textit{open-book} QA~\cite{chen2017reading, lewis2020retrieval,izacard2021leveraging, lazaridou2022internet}. 

\subsection{Retrieved vs. Generated Context}
% Therefore, we investigate using the retrieved context in the few-shot prompting setting for CBQA, and compare its performances with using generated context. 

In CBQA setting, we are not allowed to retrieve context from external knowledge sources. 
However, we can retrieve the contexts from the \sudan{supporting repository} based on their relevance to the given question. 
We use $c_r=\{c^1_r, c^2_r,..., c^m_r\}$ to represent the top-$m$ relevant context for question $Q$. 
It can be obtained via Equation~\ref{eqa:retriever_score}.
%, where $c^j_r$ is the context of the $j$-th samples $S_j$ in $S=\{S_1, S_2,... S_m\}$. 

Let the top-1 retrieved context be $c^{\text{top-1}}_r$ for question $Q$.
We use $c^{\text{top-1}}_r$ to compare with the generated context, $c_{gen}$. 
We use the same top-$m$ prompts $S'$ for answer prediction as introduced in Section~\ref{sec:ans_pred}. 
The answer $a^r_p$ for the $c^{\text{top-1}}_r$ will be:
\begin{equation}
    a^r_p = \mathcal{LM}(Prompt(\bm{c^{\text{top-1}}_r}, Q)))
    \label{equ:c1}
\end{equation}

where $Prompt(c^{\text{top-1}}_r,Q)$ can be obtained via Equation~\ref{equ:cgen_q}. 

% Equation~\ref{equ:answer_prediciton} is used to calculated the answer $A_p$ using generated context $C_{gen}$ in CGAP approach. 

% We experimented using different context generation and answer prediction LM sizes, to compare $C_{gen}$ with $C^{\text{top-1}}_r$ thoroughly. 
% We show that the accuracy gains from $C_{gen}$ are largely correlated with LM size and the the model size for answer prediction (AP) in XX.

The comparison between $c^{\text{top-1}}_r$ and $c_{gen}$ is shown in Table~\ref{tab:further_analysis_c1}. From the upper part of the table, we see that using $c^{\text{top-1}}_r$ gives slightly higher EM score than using $c_{gen}$ generated by 357M and 1.3B LMs. However, $c_{gen}$ gives higher EM scores than $c^{\text{top-1}}_r$ on all three datasets when we scale up the context generation LM size to 530B. This suggests the use of large pretrained LM for a better generated context.

\begin{table*}[!t]
\centering
\begin{adjustbox}{width=0.99\textwidth}
{
\begin{tabular}{ll}
\specialrule{.08em}{.1em}{.1em} 
\multicolumn{2}{l}{\textbf{Question}: Which sitcom star appeared on the big screening 'The Object of My Affection'?} \\ 
\textbf{Golden Answer}: [Jennifer Anniston, Jen Aniston,  ...] \\
\hline

Predicted Answer (w/o $c_{gen}$): & \textcolor{mypink3}{Ross Hatley} \\
Predicted Answer ($c^{\text{top-1}}_r$): & \textcolor{mypink3}{Laurie Metcalfe} \\
Predicted Answer ($c_{gen}$): & \textcolor{mygreen1}{Jennifer Aniston} / \textcolor{mypink3}{ Paul Rudd} /\textcolor{mypink3}{ Christine Baranski} / \textcolor{mypink3}{Lisa Kudrow} \\
Predicted Answer (($c^1_{gen}$,..., $c^k_{gen}$)): & \textcolor{mygreen1}{Jennifer Aniston} \\ 
\specialrule{.08em}{.1em}{.1em} 
\end{tabular}
}
\end{adjustbox}
\caption{Comparison of answers predicted \textit{w/o} and \textit{w/} different context. Example from TriviaQA~\cite{joshi2017triviaqa} test set. Red and green colors denote \textcolor{mypink3}{in-correct} and \textcolor{mygreen1}{correct} answer, respectively.}
\label{tab:further-example-TQA}. 
\end{table*}

\begin{table}[]
\centering
\begin{adjustbox}{width=0.46\textwidth}
{
\begin{tabular}{lc|ccc}
\specialrule{.08em}{.1em}{.1em} 
\multicolumn{1}{c}{\textbf{AP}}          & \textbf{Context}          & \textbf{NQ}    & \multicolumn{1}{l}{\textbf{TQA}} & \multicolumn{1}{l}{\textbf{WQ}} \\ \hline
\multirow{4}{*}{357M}  & $c^{\text{top-1}}_r$  & 25.1 & 32.2 & 28.3 \\
                      & $c_{gen}$ \small{(357M LM)}  & 22.9  & 28.5  & 26.2 \\
                      & $c_{gen}$ \small{(1.3B LM)}  & 23.1  & 29.7  & 28.3 \\
                      & $c_{gen}$ \small{(530B LM)}  & \textbf{26.3}  & \textbf{36.3} & \textbf{31.2}  \\ \hline
% \multirow{4}{*}{1.3B} & $C^1_r$   & 28.4 & 41.0                 & 34.5 \\
%                       & $C_{gen}$ \small{(357M LM)}  & 25.4 & 34.3 & 30.1\\
%                       & $C_{gen}$ \small{(1.3B LM)} & 26.6 & 35.0  & 31.9 \\
%                       & $C_{gen}$ \small{(530B LM)} & \textbf{31.4} & \textbf{47.0} & \textbf{37.3} \\ \hline
\multicolumn{1}{c}{\multirow{2}{*}{530B}} & $c^{\text{top-1}}_r$  & \textbf{30.8} & \textbf{58.1}  & \textbf{29.5} \\
\multicolumn{1}{c}{}  & $c_{gen}$\small{(530B LM)}  & 29.5 & 56.3  & 28.3 \\ 
\specialrule{.08em}{.1em}{.1em} 
\end{tabular}
}
\end{adjustbox}
\caption{Comparison of using retrieved top-1 context $c^{\text{top-1}}_r$, with few-shot generated context $c_{gen}$ on \textit{closed-book} QA task.}

\label{tab:further_analysis_c1}
\end{table}

\subsection{Multiple Retrievals vs. Context Marginalization}
We notice that in Table~\ref{tab:further_analysis_c1}, $c^{\text{top-1}}_r$ performs slightly better than $c_{gen}$ when using 530B LM for answer prediction. 
We argue that this might be caused by the hallucination in $c_{gen}$. 
While we have shown in Section~\ref{sec:margin} that context marginalization could mitigate the problem and improve answer accuracy, we further facilitate $c_{gen}(530B)$ with context marginalization and compare with retrieved context. 

For fair comparison, we perform majority voting using the top-$k$ retrieved context $c_r$, since ~\citet{karpukhin2020dense} showed that the quality of the retrieved documents will also affect the final answer accuracy. 
Specifically, we replace $c^{\text{top-1}}_r$ with each retrieved context $c^i_r$ in Equation~\ref{equ:c1} to predict answer $a^{r(i)}_p$ ($i=1,...,k$), and use Equation~\ref{equ:marginalization} to select the most frequent answer as the final answer.
 
Furthermore, we replace $c^{\text{top-1}}_r$ with golden context $c_{golden}$ in Equation~\ref{equ:c1}. 
This will be the upper-bound of using retrieved/generated context in the two-stage, few-shot prompting CBQA task.

% on all different model size combinations 
\begin{table}[!th]
\centering
\begin{adjustbox}{width=0.46\textwidth}{
\begin{tabular}{cl|ccc}
\specialrule{.08em}{.1em}{.1em} 
% \hline
\textbf{AP} & \multicolumn{1}{c}{\textbf{Context}} & \textbf{NQ} & \textbf{TQA} & \textbf{WQ} \\ \hline
% \multirow{5}{*}{357M} & $c_{golden}$ & \textbf{31.6} & \textbf{41.5} & 33.4 \\
%  & $c^{\text{top-1}}_r$ & 25.1 & 32.2 & 28.3 \\
%  & ($c^1_r$,...,$c^k_r$) & 24.6 & 32.4 & 27.8 \\
% & $c_{gen}$  & 26.3 & 36.3 & 31.2 \\
%  & ($c^1_{gen}$,..., $c^k_{gen}$) & 28.9 & 45.7 & 34.0 \\ \hline
%  &  &  &  &  \\
% \multirow{5}{*}{1.3B} & $c_{golden}$ & \textbf{35.1} & 51.4 & 38.1 \\
%  & $c^{\text{top-1}}_r$ & 28.4 & 41.0 & 34.5 \\
%  & ($c^1_r$,...,$c^k_r$) & 28.1 & 42.9 & 34.5 \\
%   & $c_{gen}$   & 31.4 & 47.0 & 37.3 \\
%  & ($c^1_{gen}$,..., $c^k_{gen}$) & 33.5 & \textbf{55.5} & \textbf{41.5} \\ \hline
%  &  &  &  &  \\
\multirow{5}{*}{530B} & $c_{golden}$ & 36.0 & 61.3 & 30.2 \\
 & $c^{\text{top-1}}_r$ & 30.8 & 58.1 & 29.5 \\
 & ($c^1_r$,...,$c^k_r$) & 29.5 & 56.3 & 28.3 \\
 & $c_{gen}$   & 23.0 & 55.3 & 23.6 \\
 & ($c^1_{gen}$,..., $c^k_{gen}$) & \textbf{42.0} & \textbf{68.6} & \textbf{41.8} \\ 
 \specialrule{.08em}{.1em}{.1em} 
\end{tabular}
}
\end{adjustbox}
\caption{Comparison of using context marginalization ($c^1_{gen}$,..., $c^k_{gen}$), multiple retrievals ($c^1_r$,...,$c^k_r$), and golden context $c_{golden}$ on \textit{closed-book} QA task.}
% \vspace{-15pt}
\label{tab:further_analysis_margin}
\end{table}

We show the results in Table~\ref{tab:further_analysis_margin}. 
As we can see, using marginalization over $c_{gen}$ consistently outperforms $c^{\text{top-1}}_r$, and also better than majority voting over multiple retrieved contexts $c_r$ for answer prediction 
% \sudan{on different model size combinations} 
on all three datasets. Notably, marginalization over $c_{gen}$ yields higher EM score than using $c_{golden}$ when using 530B LM for answer prediction.
\sudan{We observed similar trends when experimented on 357M and 1.3B parameter models. In Table~\ref{tab:further-example-TQA}, we show a concrete example that compares using different context for answer generation for better understanding\footnote{More concrete comparison examples are shown in Appendix~\ref{sec:appendix_example} Table~\ref{tab:example-NQ} and Table~\ref{tab:example-WQ}.}.}
% The possible reason for this might be for some $C_{golden}$ that can not lead to a correct answer, the $\{c^1_{gen},..., c^k_{gen}\}$ generated by the 530B model can suprisingly .  

\section{Related Works}

\paragraph{Open-domain QA} is the task of answering general-domain questions~\cite{chen2017reading}, in which the evidence is usually not given. Models that explicitly exploit an external corpus are referred as \textit{open-book} models~\cite{roberts2020much}. They typically index the corpus and then \textit{retrieve-and-read} to extract the answer span from documents~\cite{chen2017reading, lee2019latent, izacard2021leveraging, rag, lazaridou2022internet, su2022read}. Another recently proposed class of methods is \textit{closed-book} QA models. ~\citet{ye2020studying,roberts2020much} finetune pretrained LMs such as T5~\cite{raffel2020exploring} or BART~\cite{lewis2020bart} with QA pairs without access to any external knowledge or context.
 
\paragraph{Few-shot LM Prompting} ~\citet{radford2019language, brown2020language} prompt GPT-2~\cite{radford2019language} and GPT-3~\cite{brown2020language} conditioned on several few-shot examples to predict the answer for ODQA. Most recent work by ~\citet{lazaridou2022internet} further empower LM's few-shot prompting abilities with information returned from the web using Google-Search API, and experimented on QA task. While ~\citet{wei2022chain, wang2022self} use \textit{chain of thought} few-shot prompting of LM to generate a coherent chain of short sentences that minic the reasoning process of human might employ to solve reasoning tasks.  

\section{Conclusion}
We propose a simple yet effective framework named \textbf{CGAP} for open-domain QA. CGAP performs \textbf{C}ontext \textbf{G}eneration followed by  \textbf{A}nswer \textbf{P}rediction via two-stage prompting using large pretrained LMs. 
It does not rely on external knowledge sources, and does not need finetuning or add extra learnable parameters.
To the best of our knowledge, we are the first to leverage generated context from large pretrained LMs for open-domain QA.
Experimental results on three QA benchmarks show that our method significantly outperforms previous \textit{closed-book} QA methods and is par with \textit{open-book} methods. We demonstrate our method up to 530B parameter models and showcase that larger models boost the accuracy by huge margins.

\section{Limitations}
% As we show in the paper, \textbf{CGAP} has obtained satisfactory results on \textit{closed-book} QA task. However, the method have some limitations. First, the performance of CGAP will be affected by the size of LMs it uses, as we shown in Figure~\ref{Fig:ablation-CG-357m-1.3b-ans}. In Section~\ref{sec:main_results}, our highest accuracy results reported in Table~\ref{tab:main_results} was based on gigantic pretrained LMs sized 530B, which is only accessible via API. Second, as we show in Section~\ref{sec:margin}, using context marginalization with several pretrained LMs can also yield satisfied results. However, as Equation~\ref{equ:marginalization} shows, the context marginalization relies on sampling a LM $k$ times, or querying $k$ LMs in parallel. While normally it requires one to multiple GPUs to run large pretrained LM, using $k$ LMs at the same time will be even more demanding for GPU resources.
\sudan{
As we show in the paper, \textbf{CGAP} has obtained satisfactory results on open-domain QA task. However, the method have limitations. The accuracy of CGAP will be affected by the size of LMs it uses, as we shown in Figure~\ref{Fig:ablation-CG-357m-1.3b-ans}. In Section~\ref{sec:main_results}, our highest accuracy results reported in Table~\ref{tab:main_results} used a large 530B pretrained LM, which is only accessible via API. Also, the generated context may contain hallucinated content.}

% (Third, the generated context ...)

% While we are open to different types of limitations, just mentioning that a set of results have been shown for English only probably does not reflect what we expect. Mentioning that the method works mostly for languages with limited morphology, like English, is a much better alternative. In addition, limitations such as low scalability to long text, the requirement of large GPU resources, or other things that inspire crucial further investigation are welcome.

\bibliography{anthology,custom}
\bibliographystyle{acl_natbib}

\appendix

\section{Standard Few-shot Prompting of GPT-3}
\label{sec:appendix-gpt3}
~\citet{brown2020language} adopted the standard few-shot prompting on GPT-3, and evaluated on the three open-domain QA datasets NQ~\cite{kwiatkowski2019natural}, WQ~\cite{berant2013semantic} and TQA~\cite{joshi2017triviaqa}, for \textit{closed-book} QA task. In order to compare with their reported results, we re-implement their method using the same few-shot configuration as described in the paper and query the OpenAI API.

\paragraph{Experimental Setups} As OpenAI hasn’t officially release information about their API model sizes, we deduce the sizes of OpenAI API models based on their performances from EleutherAI's blog\footnote{https://blog.eleuther.ai/gpt3-model-sizes/}. Specifically, we query Ada and Babbage models' API, trying to reproduce the reported results for GPT-3 Medium (350M) and GPT-3 XL (1.3B) models, respectively.

We use two prompt formats to query the OpenAI API. The first prompt format is the one described in the paper~\cite{brown2020language} (referred as \textit{GPT-3 format}): randomly draw 64 question-answer pairs from the corresponding \sudan{supporting repository}, and use 'Q: ' and 'A: ' respectively as prefix before each question and answer, to build the conditioning prompts. We also use the prompt format from EleutherAI's language model evaluation harness github\footnote{https://github.com/EleutherAI/lm-evaluation-harness} (referred as \textit{EleutherAI}). Furthermore, we experiment using the same prompting format as we used in our standard prompting baseline (LM-530B) in Section~\ref{sec:baselines} (referred as \textit{Our format}), and prompting the LM of size 357M and 1.3B to compare.

\paragraph{Results} We show the results of prompting GPT-3 under zero-shot, one-shot and few-shot settings in Table~\ref{tab:appendix-gpt-3-zeroshot}, Table~\ref{tab:appendix-gpt3-oneshot} and Table~\ref{tab:appendix-gpt3-few-shot} respectively. As we can see, no matter what prompting formats we use, the results reported in the GPT-3 paper~\cite{brown2020language} are almost always higher than our reproduced ones on all three datasets, over the two different LM sizes. The gaps become even larger at few-shot setting. Thus we conjuncture that we are not able to reproduce the results reported by ~\citet{brown2020language} using GPT-3 (175B) on the three QA datasets. So we did not include their reported results to compare with our CGAP method in Table~\ref{tab:main_results}.

Furthermore, we notice that the results based on our baseline's prompting configuration are always on par with the results from querying OpenAI API. Thus we believe that the \textbf{LM-530B} is a reliable and fair standard few-shot prompting baseline to compare with.

\section{Examples}
\label{sec:appendix_example}
We show three examples from NQ and WQ test set in Table~\ref{tab:example-NQ} and Table~\ref{tab:example-WQ} respectively. In each table, we show the predicted answers from (1) standard prompting, (2) two-stage prompting using top-1 retrieved context $c^{\text{top-1}}_r$, (3) CGAP w/o marginalization, and (4) CGAP. All those predicted answers are based on LMs of size 530B.

We also show an example illustrate CGAP with 8 generated context and their corresponding predicted answer in Table~\ref{tab:appendix-gcap}. As we can see, the contexts that contains lot of factually inaccurate or irrelevant content (e.g. generated context 1, 2, 4, 5, 8), thus the corresponding answer is wrong/inaccurate. However, the context generation LM also generates contexts that are more relevant and factual (e.g. generated context 3, 6, 7), and they help the answer prediction LM generate a correct answer. Therefore, CGAP can predict the final answer correctly based on marginalization over generated contexts.

\begin{table*}[]
\centering
\begin{adjustbox}{width=\textwidth}{}
{
\begin{tabular}{ccllccc}
\hline
\multicolumn{2}{c}{\multirow{2}{*}{\textbf{Model Sizes}}} & \multirow{2}{*}{\textbf{Model sources}} & \multicolumn{1}{c}{\multirow{2}{*}{\textbf{Prompting format}}} & \multicolumn{3}{c}{\textbf{zero-shot}} \\ \cline{5-7} 
\multicolumn{2}{c}{} &  & \multicolumn{1}{c}{} & \textbf{NaturalQuestion} & \textbf{TriviaQA} & \multicolumn{1}{l}{\textbf{WebQuestion}} \\ \hline
\multicolumn{2}{c}{} & GPT-3 Medium & GPT-3 paper~\cite{brown2020language} & \textbf{1.75} & \textbf{7.61} & \textbf{3.20} \\
\multicolumn{2}{c}{\multirow{4}{*}{350M}} & \multirow{2}{*}{OpenAI API (Ada)} & GPT-3  format & 1.36 & 5.45 & 1.92 \\
\multicolumn{2}{c}{} &  & EleutherAI & 1.39 & 5.54 & 2.46 \\
\multicolumn{2}{c}{} & LM-357M & Our format & 1.41 & 5.04 & 2.12 \\ \hline
\multicolumn{2}{c}{} & GPT-3 XL & GPT-3 paper ~\cite{brown2020language} & \textbf{4.40} & \textbf{19.70} & 4.63 \\
\multicolumn{2}{c}{\multirow{4}{*}{1.3B}} & \multirow{2}{*}{OpenAI API (Babbage)} & GPT-3  format & 2.27 & 9.84 & 2.12 \\
\multicolumn{2}{c}{} &  & EleutherAI & 2.47 & 12.77 & 5.22 \\
\multicolumn{2}{c}{} & LM-1.3B & Our format & 3.88 & 14.13 & \textbf{5.61} \\ \hline
\end{tabular}
}
\end{adjustbox}
\caption{Standard zero-shot prompting of GPT-3 for open-domain QA. }
\label{tab:appendix-gpt-3-zeroshot}
\end{table*}

\begin{table*}[]
\centering
\begin{adjustbox}{width=\textwidth}{}
{
\begin{tabular}{ccllccc}
\hline
\multicolumn{2}{c}{\multirow{2}{*}{\textbf{Model Sizes}}} & \multirow{2}{*}{\textbf{Model sources}} & \multicolumn{1}{c}{\multirow{2}{*}{\textbf{Prompting format}}} & \multicolumn{3}{c}{\textbf{one-shot(k=1)}} \\ \cline{5-7} 
\multicolumn{2}{c}{} &  & \multicolumn{1}{c}{} & \textbf{NaturalQuestion} & \textbf{TriviaQA} & \multicolumn{1}{l}{\textbf{WebQuestion}} \\ \hline
\multicolumn{2}{c}{} & GPT-3 Medium & GPT-3 paper~\cite{brown2020language} & \textbf{3.07} & \textbf{12.90} & \textbf{6.20} \\
\multicolumn{2}{c}{\multirow{4}{*}{350M}} & \multirow{2}{*}{OpenAI API (Ada)} & GPT-3  format & 1.83 & 10.26 & 5.07 \\
\multicolumn{2}{c}{} &  & EleutherAI & 1.77 & 10.02 & 5.61 \\
\multicolumn{2}{c}{} & LM-357M & Our format & 2.24 & 9.75 & 5.12 \\ \hline
\multicolumn{2}{c}{} & GPT-3 XL & GPT-3 paper~\cite{brown2020language} & \textbf{5.43} & \textbf{26.50} & 9.15 \\
\multicolumn{2}{c}{\multirow{4}{*}{1.3B}} & \multirow{2}{*}{OpenAI API (Babbage)} & GPT-3  format & 3.55 & 20.56 & 8.27 \\
\multicolumn{2}{c}{} &  & EleutherAI & 3.55 & 21.45 & \textbf{9.45} \\
\multicolumn{2}{c}{} & LM-1.3B & Our format & 4.71 & 21.21 & 8.76 \\ \hline
\end{tabular}
}
\end{adjustbox}
\caption{Standard one-shot prompting of GPT-3 for open-domain QA. }
\label{tab:appendix-gpt3-oneshot}
\end{table*}

\begin{table*}[]
\centering
\begin{adjustbox}{width=\textwidth}{}
{
\begin{tabular}{ccllccc}
\hline
\multicolumn{2}{c}{\multirow{2}{*}{\textbf{Model Sizes}}} & \multirow{2}{*}{\textbf{Model sources}} & \multicolumn{1}{c}{\multirow{2}{*}{\textbf{Prompting format}}} & \multicolumn{3}{c}{\textbf{few-shot(k=64)}} \\ \cline{5-7} 
\multicolumn{2}{c}{} &  & \multicolumn{1}{c}{} & \textbf{NaturalQuestion} & \textbf{TriviaQA} & \multicolumn{1}{l}{\textbf{WebQuestion}} \\ \hline
\multicolumn{2}{c}{} & GPT-3 Medium & GPT-3 paper~\cite{brown2020language} & \textbf{4.46} & \textbf{16.30} & \textbf{12.60} \\
\multicolumn{2}{c}{\multirow{4}{*}{350M}} & \multirow{2}{*}{OpenAI API (Ada)} & GPT-3  format & 3.43 & 12.46 & 10.73 \\
\multicolumn{2}{c}{} &  & EleutherAI & 3.71 & 12.46 & 10.29 \\
\multicolumn{2}{c}{} & LM-357M & Our format & 3.85 & 11.66 & 10.97 \\ \hline
\multicolumn{2}{c}{} & GPT-3 XL & GPT-3 paper~\cite{brown2020language} & \textbf{9.72} & \textbf{32.10} & \textbf{19.60} \\
\multicolumn{2}{c}{\multirow{4}{*}{1.3B}} & \multirow{2}{*}{OpenAI API (Babbage)} & GPT-3  format & 8.28 & 24.70 & 18.95 \\
\multicolumn{2}{c}{} &  & EleutherAI & 7.81 & 24.93 & 18.16 \\
\multicolumn{2}{c}{} & LM-1.3B & Our format & 7.87 & 24.88 & 17.52 \\ \hline
\end{tabular}
}
\end{adjustbox}
\caption{Standard few-shot (k=64) prompting of GPT-3 for open-domain QA.}
\label{tab:appendix-gpt3-few-shot}
\end{table*}

% They use beam search with a beam width of 4 and a length penalty of 0.6 for answer prediction. 

\clearpage
\begin{table*}[]
\centering
\begin{adjustbox}{width=0.95\textwidth}
{
\begin{tabular}{ll}
\hline
\multicolumn{2}{l}{\textbf{Question}: When is the next deadpool movie being released?} \\ \hline
Golden Answer: & {[}May 18, 2018{]} \\
Predicted Answer (standard prompting): & \textcolor{mypink3}{Prime availability TBD} \\
Predicted Answer ($c^{\text{top-1}}_r$): & \textcolor{mygreen1}{May 18, 2018} \\
Predicted Answer (CGAP w/o marginalization): & \textcolor{mygreen1}{May 18, 2018} / \textcolor{mypink3}{date21-May-2018} / \textcolor{mypink3}{May 29, 2019} /\textcolor{mypink3}{16th May 2018} \\
Predicted Answer (CGAP): &\textcolor{mygreen1}{May 18, 2018} \\ \hline
\end{tabular}
}
\end{adjustbox}
\caption{Example from NQ~\cite{kwiatkowski2019natural} test set. Red and green colors denote \textcolor{mypink3}{in-correct} and \textcolor{mygreen1}{correct} answer, respectively.}
\label{tab:example-NQ}
\end{table*}

% \footnote{For predicted answer from CGAP w/o marginalization, we show all the different answers from 8.}

% \begin{table*}[]
% \centering
% \begin{adjustbox}{width=\textwidth}
% {
% \begin{tabular}{ll}
% \hline
% \multicolumn{2}{l}{\textbf{Question}: Which sitcom star appeared on the big screening 'The Object of My Affection'?} \\ \hline
% Golden Answer: & [Jennifer Anniston, Jen Aniston,  ...] \\
% Predicted Answer (standard prompting): & \textcolor{mypink3}{Ross Hatley} \\
% Predicted Answer ($c^{\text{top-1}}_r$): & \textcolor{mypink3}{Laurie Metcalfe} \\
% Predicted Answer (CGAP w/o marginalization): & \textcolor{mygreen1}{Jennifer Aniston} / \textcolor{mypink3}{ Paul Rudd} /\textcolor{mypink3}{ Christine Baranski} / \textcolor{mypink3}{Lisa Kudrow} \\
% Predicted Answer (CGAP): & \textcolor{mygreen1}{Jennifer Aniston} \\ \hline
% \end{tabular}
% }
% \end{adjustbox}
% \caption{Example from TriviaQA~\cite{joshi2017triviaqa} test set.}
% \label{tab:example-TQA}
% \end{table*}

\begin{table*}[]
\centering
\begin{adjustbox}{width=\textwidth}
{
\begin{tabular}{ll}
\hline
\multicolumn{2}{l}{\textbf{Question}: Where George Lopez was born?} \\ \hline
Golden Answer: & {[}Mission Hills{]} \\
Predicted Answer (standard prompting): & \textcolor{mypink3}{San Fernando, California} \\
Predicted Answer ($c^{\text{top-1}}_r$): & \textcolor{mypink3}{Washington Heights} \\
Predicted Answer (CGAP w/o marginalization): & \textcolor{mygreen1}{Mission Hills} / \textcolor{mypink3}{Los Angeles}/ \textcolor{mypink3}{San Fernando} /\textcolor{mypink3}{ Castle Hill} /\textcolor{mypink3}{ canada }\\
Predicted Answer (CGAP): & \textcolor{mygreen1}{Mission Hills} \\ \hline
\end{tabular}
}
\end{adjustbox}
\caption{Example from WebQuestions~\cite{berant2013semantic} test set.}
\label{tab:example-WQ}
\end{table*}

\clearpage

\begin{table*}[]
\small
\centering
\begin{adjustbox}{width=\textwidth}
{
\begin{tabularx}{\textwidth}{X}
\hline
\textbf{Question}: Where George Lopez was born?       \textbf{Golden Answer:} [Mission Hills]\\
\hline
\\
\textbf{Generated Context 1}: George Lopez They are American citizens, born in Los Angeles, to Gregorio Lopez (October 26, 1934 – January 25, 2011) and Maria Torres (October 12, 1938 – October 25, 2014), Mexican immigrants who were born in Reynosa, Tamaulipas. Lopez's paternal grandparents were Mexican, while his maternal grandparents were of Mexican and Spanish descent. Lopez's father was a worker for the Los Angeles Metropolitan Transit Authority. He married Maria in Los Angeles when he was 16 years old, and she was 15. Maria filed for divorce from Gregorio in the mid-1970s. \\ 
\textbf{Predicted Answer:} Los Angeles \\
\\
\textbf{Generated Context 2}: George Lopez born April 23, 1961. The family eventually moved from Marín to Los Angeles in the mid-1960s. After Lopez graduated from San Fernando High School in 1979, he attended the California State University, Los Angeles. It was there where Lopez discovered his comedic talent after a video of him doing a stand-up routine made it into the hands of "Tonight Show" booker, Fred De Cordova, in 1985. \\
\textbf{Predicted Answer:} canada \\
\\
\textbf{Generated Context 3}: George Lopez at 20th Century Fox Television until 2010, when he began appearing in his self-produced sitcom, George Lopez. Lopez was born on April 23, 1961, in Mission Hills, Los Angeles, California. He was raised by his maternal grandmother, Benita Gutierrez, a factory worker, and his maternal grandfather, Refugio Gutierrez, who worked as a janitor and metal fabricator at an ice skating rink. \\
\textbf{Predicted Answer:} Mission Hills \\
\\
\textbf{Generated Context 4}: George Lopez in San Fernando, California. He is of Mestizo Mexican descent. His father, Saul Lopez, was from Sonora, and his mother, Elvira, was a native of Arizona. They were migrant workers who picked crops and moved often, sometimes taking their seven children with them. While growing up, Lopez did not speak English until he was seven. Lopez attended Our Lady of Lourdes, a Roman Catholic elementary school in Pico Rivera. \\
\textbf{Predicted Answer:} San Fernando \\
\\
\textbf{Generated Context 5}: George Lopez A theater called "The Tower", which now houses a branch of Bank of America, stands on the former site of his childhood home. Lopez was born at Daniel Freeman Memorial Hospital in Inglewood, California and was raised by his maternal grandmother, Benita Gutierrez, a factory worker who took him to the movies every week. He grew up in the Echo Park neighborhood of Los Angeles, California and attended many different schools in the Los Angeles Unified School District.\\
\textbf{Predicted Answer:} Los Angeles \\
\\
\textbf{Generated Context 6}: George Lopez López was born in Mission Hills, Los Angeles, California. He is the son of Frieda and Anatasio López, Mexican immigrants from Tepic, Nayarit, Mexico. Lopez's father was a farm worker, and his mother worked at a canning factory. Lopez has stated that he had a difficult childhood; his parents divorced when he was five, and his mother struggled to raise him and his two sisters. \\
\textbf{Predicted Answer:}  Mission Hills \\
\\
\textbf{Generated Context 7}: George Edward Lopez. Lopez was born in Mission Hills, Los Angeles. His parents, Frieda and Anatasio Lopez, were both born and raised in Texas. His mother was of Mexican descent and his father, also called ""Charlie"", was of Sicilian descent. Lopez grew up with two older sisters. He attended San Fernando High School in the Los Angeles neighborhood of San Fernando.\\
\textbf{Predicted Answer:} Mission Hills \\
\\
\textbf{Generated Context 8}: George Lopez had in his mouth. He was raised by his mother, Frieda, a telephone company employee, and his grandmother, Benita Gutierrez, a factory worker. Lopez graduated from the Roman Catholic Saint Monica's High School in 1974, where he met his future wife, Ann Serrano. He moved to California and graduated from the California State University, Los Angeles, in 1986 with a degree in communications.\\
\textbf{Predicted Answer:} Castle Hill \\ 
\\\hline
\textbf{Predicted Answer (CGAP):}  Mission Hills \\

\hline
\end{tabularx}
}
\end{adjustbox}
\caption{Example illustrates CGAP, in which we marginalize over 8 generated contexts.}
\label{tab:appendix-gcap}
\end{table*}

\begin{table*}[]
\small
\centering
\begin{adjustbox}{width=\textwidth}
{
\begin{tabularx}{\textwidth}{X}
\hline

\\
$Q$: Who was the first U.S. president to be born in the United States of America? \\
$A$: with the issue of 1863 (displayed in First appearances) above). At that time, as aforesaid, he had already appeared on two different Confederate 200a2 stamps. Andrew Jackson appears on the following commemorative issues: Martin Van Buren (December 5, 1782 - 2013 July 24, 1862) was the eighth President of the United States from 1837 to 1841. Before his presidency, he served as the eighth Vice President (1833~20131837) and the 10th Secretary of State under Andrew Jackson. Van Buren was the first U.S. President to be born an American citizen. William Henry Harrison (February 9, 1773 \u2013 April 4, 1841) was the \\

$Q$: Which US President who was in office 1923-29, was the only one to be born on 4th July? \\
$A$: Calvin Coolidge John Calvin Coolidge Jr. (; July 4, 1872 2013 January 5, 1933) was an American politician and the 30th President of the United States from 1923 to 1929. A Republican lawyer from New England, born in Vermont, Coolidge worked his way up the ladder of Massachusetts state politics, eventually becoming governor. His response to the Boston Police Strike of 1919 thrust him into the national spotlight and gave him a reputation as a man of decisive action. Soon after, he was elected Vice President of the United States in 1920, and succeeded to the presidency upon the sudden \\
$Q$: What long running comic strip, introduced on Oct 2, 1950, was drawn by the same cartoonist, Charles M. Schultz, during its' entire run? \\
$A$: Peanuts Peanuts is a syndicated daily and Sunday American comic strip written and illustrated by Charles M. Schulz that ran from October 2, 1950, to February 13, 2000, continuing in reruns afterward. "Peanuts" is among the most popular and influential in the history of comic strips, with 17,897 strips published in all, making it "arguably the longest story ever told by one human being. At its peak in the mid-to late 1960s, "Peanuts" ran in over 2,600 newspapers, with a readership of around 355 million in 75 countries, and was translated into 21 languages. It helped to cement the" \\
$Q$: In the comic strip "Peanuts", whose father is a barber? \\
$A$: Franklin (Peanuts) Franklin Armstrong is a character in the long-running comic strip "Peanuts", created by Charles M. Schulz. Introduced on July 31, 1968, Franklin was the first African American character in the strip. He goes to school with Peppermint Patty and Marcie. In his first story arc, he met Charlie Brown when they were both at the beach. Franklin's father was a soldier fighting in Vietnam, to which Charlie Brown replied "My dad's a barber...he was in a war too, but I don't know which one." Franklin later paid Charlie Brown a visit and found some of Charlie Brown's other \\
$Q$: Which US President was given the nickname 'The Trust Buster'? \\
$A$: Howard Taft. A Progressive reformer, Roosevelt earned a reputation as a "trust buster" through his regulatory reforms and anti-trust prosecutions. His presidency saw the passage of the Pure Food and Drug Act, which established the Food and Drug Administration to regulate food safety, and the Hepburn Act, which increased the regulatory power of the Interstate Commerce Commission. Roosevelt took care, however, to show that he did not disagree with trusts and capitalism in principle, but was only against monopolistic practices. His "Square Deal" included regulation of railroad rates and pure foods and drugs; he saw it as a fair deal \\
$Q$: American cartoonist Thomas Nast is credited with creating the traditional image of which popular Christmas character? \\
$A$: Thomas Nast Thomas Nast (September 27, 1840 ~ December 7, 1902) was a German-born American caricaturist and editorial cartoonist considered to be the "Father of the American Cartoon". He was the scourge of Democratic Representative "Boss" Tweed and the Tammany Hall Democratic party political machine. Among his notable works were the creation of the modern version of Santa Claus (based on the traditional German figures of Sankt Nikolaus and Weihnachtsmann) and the political symbol of the elephant for the Republican Party (GOP). Contrary to popular belief, Nast did not create Uncle Sam (the male personification of the United \\
$Q$: October 2, 1950 saw the debut of what comic strip which ran until creator Charles M. Schulz's death on Feb 12, 2000? \\
$A$: from a wide range of acclaimed contemporary artists and designers who have been inspired by the cartoon. Peanuts Peanuts is a syndicated daily and Sunday American comic strip written and illustrated by Charles M. Schulz that ran from October 2, 1950, to February 13, 2000, continuing in reruns afterward. "Peanuts" is among the most popular and influential in the history of comic strips, with 17,897 strips published in all, making it "arguably the longest story ever told by one human being". At its peak in the mid- to late 1960s, "Peanuts" ran in over 2,600 newspapers, with a readership of \\
$Q$: In which state was Charles Schulz born? \\ 
$A$: Charles M. Schulz Charles Monroe "Sparky" Schulz (November 26, 1922 ~ February 12, 2000), nicknamed Sparky, was an American cartoonist. Schulz is known for the comic strip "Peanuts" (which featured the characters Charlie Brown and Snoopy, among others). He is widely regarded as one of the most influential cartoonists of all time, cited by cartoonists including Jim Davis, Bill Watterson, and Matt Groening. Born in Minneapolis, Minnesota, Schulz grew up in Saint Paul. He was the only child of Carl Schulz, who was born in Germany, and Dena Halverson, who had Norwegian heritage. His uncle called him "Sparky" after \\
$Q$: \textbf{Who was President when the first Peanuts cartoon was published?} \\
\\\hline

\hline
\end{tabularx}
}
\end{adjustbox}

\caption{$Prompt(Q)$ Example. For the question \textbf{"Who was President when the first Peanuts cartoon was published?"} from TQA~\cite{joshi2017triviaqa}, we selected 8 $<q_i, c_i>$ samples from the supporting repository $\mathcal{D}$, and construct the $Prompt(Q)$ as above. to prompt LMs for $c_{gen}$ generation.}
\label{tab:appendix-promptq}
\end{table*}
% % Entries for the entire Anthology, followed by custom entries
% \bibliography{anthology,custom}
% \bibliographystyle{acl_natbib}

\end{document}